# A Framework for Semi-automated Web Service Composition in Semantic Web

Debajyoti Mukhopadhyay, Archana Chougule

*Department of Information Technology*
*Maharashtra Institute of Technology*
*Pune 411038, India*
debajyoti.mukhopadhyay@gmail.com
chouguleab@gmail.com

*Abstract*— **Number of web services available on Internet and its usage are increasing very fast. In many cases, one service is not enough to complete the business requirement; composition of web services is carried out. Autonomous composition of web services to achieve new functionality is generating considerable attention in semantic web domain. Development time and effort for new applications can be reduced with service composition. Various approaches to carry out automated composition of web services are discussed in literature. Web service composition using ontologies is one of the effective approaches. In this paper we demonstrate how the ontology based composition can be made faster for each customer. We propose a framework to provide precomposed web services to fulfil user requirements. We detail how ontology merging can be used for composition which expedites the whole process. We discuss how framework provides customer specific ontology merging and repository. We also elaborate on how merging of ontologies is carried out.**

## I. INTRODUCTION

Many enterprise applications run by using various web services available online. They provide a mechanism for integrating heterogeneous applications required for business. As number of such available services is very large, comparing a set of web services for matching with customer needs and selecting best services among them is becoming a critical task. In case, where one web service is not enough to fulfill the user requirements, composition of web services is carried out. In Semantic Web, information is represented in machine readable format, to make business processes fully automated. In case of web service composition in Semantic Web, it is carried out with help of ontological descriptions of web services. Ontology is a formal explicit description of concepts in a domain of discourse, properties of each concept describing various features and attributes of the concept, and restrictions on slots[13]. Languages like DAML and OWL-S are used to specify ontological descriptions for web services. Every time carrying out the composition at runtime delays response to customer. To overcome this problem, we need to have automated system for providing web services access. Semantic web services provide solutions to the challenges associated with automated discovery and dynamic composition tasks provided by service-based systems. We propose a framework for providing many composite services through one integrated service. Using framework, a repository of merged ontologies for composite services is maintained. With the help of this repository, customer is served immediately if request comes for the composite service. Merged ontology for requested service is already available in the repository. For designed framework, we consider that all ontological descriptions of web services are in OWL-S format.

The advantage of our approach is, the time required for serving composite web service request is lesser than traditional approach. The paper describes details about ontology based web service composition and ontology merging technique used by the framework. We demonstrate how our approach reduces overhead of customer while using composition of web services.

Rest of the paper is structured as follows: we first describe the work done by other researchers regarding ontology based web service composition in Section 2. We discuss a general method of ontology based web service composition in Section 3. In Section 4, we describe our framework for providing precomposed web services. Implementation of framework is described in Section 5. We have implemented a framework to provide web services under e-learning domain. In Section 6, we show impact analysis of implemented system. We conclude the paper in Section 7.

## II. RELATED WORK

Various approaches are suggested by researchers to carry out web service composition using domain ontologies and service ontologies. Semi-automatic web service composition using Business Process Execution Language is implemented by G. Li, S. Deng, H. Xia and Chuan Lin[1]. A process ontology tree is built using Work flow technology. To complete automated composition, all rules and relations are extracted from a process ontology tree. After that, the target nodes are searched in process ontology tree and validation of interrelations is carried out. Heuristic approach for web service composition is also suggested by YU Qing-mei, XIAO Peng-yan and JIN Ting [2], where it finds similarity based on ontology. They construct a web service composition graph. They generate a serial of composition path using directed acyclic graph. They introduce a heuristic function which reduces the searching area. The explained algorithm considers services semantic similarity and adjusts composition plan dynamically. Jiangang Ma, Yanchun Zhang and Minglu Li[3] explain goal driven and ontology based approach for web service composition. In this architecture, they decompose the user's goal to sub goals. Information in the goal and web services is annotated with domain specific ontology. For carrying out composition of web services they use AI technology and theory of reasoning about action. A two way composition method is presented by Yajuan Song and Lei Liu [4]. They process domain ontology for dynamic web service composition. Domain ontology is annotated with the input and output parameters of the web service according to the concepts in domain ontology they belong to. Static and dynamic service composition environments are suggested by Liquan Han and Shufen Liu [5]. They apply knowledge reasoning process to service ontology for composition. With the definition and application of ontology, they propose the concept and definition format of service-ontology, and with the construction of service composition and corresponding algorithm, they apply it in the dynamic Web service composition application. Combined with intelligent smart transcript repository and ontology knowledge repository, they describe the static and dynamic composition environments to facilitate knowledge accumulation, logic inference, binding web service and creation of process driven model. Approach considers semantics of services along with quality and efficiency of the composition. B. Arpinar, R. Zhang, B. Aleman-Meza and Angela Maduko [6] suggest a web service composition method which is based on ontology. They provide service profiles in DAML-S. It is used for semantic descriptions of service interfaces and functions. Naiwen Lin, Ugur Kuter and James Hendler[7] suggested web service composition based on a Hierarchical Task Network planning model. Using this planning model, they decompose the composition problem to evaluate structural properties spanning across multiple ontologies and solve the problem using existing techniques in HTN-DL. They have described the task decomposition rules based on boolean queries of terminological components. An ontology based tool called as Visualization and Formalization Tool (VFT) is implemented by C. Ma, Y. He, N. Xiong and L. Yang [8]. Service composition is carried out in two stages i.e. Logical stage and Physical stage. VFT is used at logical stage. They use VFT for visualization and formalization of web service composition. In VFT, concept ontology is loaded to build up the Service View. The output of VFT is owl file. It is then used as input for AI planner. AI planner generates the abstract workflow for composition from given owl input file and planning information. Planning and ontology concept relevance problems are combined by O. Hatzi, G. Meditskos, D. Vrakas and N. Bassiliades[9] to carry out automatic composition of web services. The system is called as PORSE II. The system parses OWL-S web service profiles and produces the web service descriptions. These descriptions are used for calculating concept similarity and determining semantically relevant concepts. Collected results are then used to represent the web service composition problem in planning terms. The solution is carried out by performing semantic relaxation and invoking external planners. Visualization and evaluation of the solution of the problem is also provided by PORSE II system. F. Tou, Q. Hu, Y. Yao, G. Xu and M. Fang [10] propose web service matching composition algorithm for bionic manufacturing domain-specific ontology. The algorithm is based on recursive principle. Heuristic knowledge of knowledge base and the semantic information of ontology base of bionic manufacturing are used for service matching and semantic reasoning. Auto-link algorithm is applied to generate tree of service composition. In that, they have searched all feasible sub-service compositions and linked together every feasible sub-service based on its inputs, outputs, preconditions and effects to construct the structured framework of service flow. The system structure of service matching and composition module is composed of four parts as request parser, service matching engine, service composability judgment processor and composed service description processor. Conceptualization of keywords is done by request parser by mapping service request keywords to domain-specific unified descriptions. Service searching engine searches services with service matching algorithm. Whether two searched services can be composed or not is decided by service composability judgment processor. At last, composed service description processor updates the description of composed service according to auto-link service composition algorithm.

## III. WEB SERVICE COMPOSITION USING ONTOLOGY

A growing number of Web Services have emerged as the Internet develops at a fast rate in recent years. The need for composing existing web services into more complex services is also increasing to support business-to-business or enterprise application integration. Because of inability of a single web service to achieve a user's goals by itself, web service composition is required. So that a collection of interacting web services can accomplish these goals. In complex examples of service compositions, so much of data collection services might be involved in a composition. In those cases, semi-automatic composition reduces time required for composition generation and execution substantially. Web service composition can be static or dynamic. AI based composition and two way composition and composition based on acyclic graphs are the examples of dynamic web service composition technique. Out of the available methods, ontology based dynamic web service composition is one of the popular fast method in the world of Semantic Web as it makes automation possible and it is more efficient. In ontology based composition, for a newly registered service, mapping to the concepts of the ontology under specific domain is carried out. Web services are classified by the concepts according to input and output parameters for functions and stored in the ontology base. Composition process is made automated by using the details specified in the domain ontology and service ontologies under related domain ontology. For each web service, service profile is written using OWL-S. It is used by Knowledge reasoner to carry out dynamic service composition. Service profile of registered web service is written using OWL-S. It is a mark-up of web services which provides a declarative, computer-interpretable API that includes the semantics of the arguments to be specified when executing the composition [11]. Service Profile of web service describes a web service as function of three basic types of information: which organization provides the service, which function is computed by the service, and a host of features that specify characteristics of the service. Domain ontology of web service is annotated after profile registration, using registered service ontology.

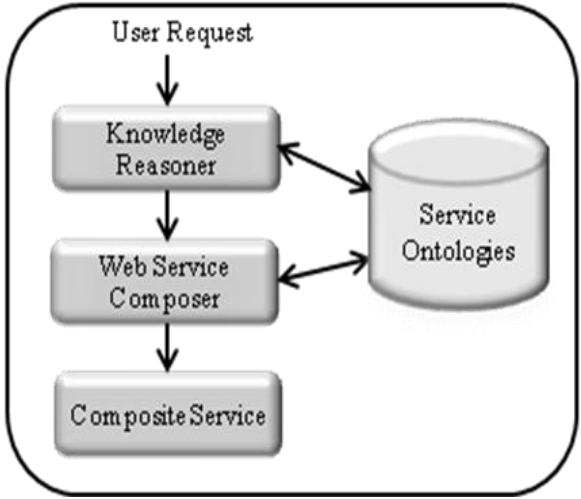

Fig. 1. Ontology based web service composition

Matching web services are found with the help of knowledge reasoner. Service matcher, sub-part of knowledge reasoner is responsible to find out matching web service from available services. It semantically searches a single web service which satisfies the user request and response is sent to user. In case, where no single web service can satisfy the user request, web service composer is used to compose a serial of web services to satisfy the request. In ontology based composition, it uses service ontologies for requirements matching. When more than two services are involved in composition, composition starts from the service that needs one or more of input parameters matching to the input parameters in the user request. If the preconditions, effects and required output parameters mentioned in user of user request are compatible to the precondition provided in composition the composition can be carried out successfully. The main advantage of this approach is, time required for searching and matching the service is less. With ontology based web service composition, one can assemble services into processes for easier and better quality workflow executions under increasing number and complexity of web services [6].

## IV. PROPOSED APPROACH

In this section, we describe our approach in detail. We first discuss about modules of the designed framework. We further describe how these modules are utilized to provide precomposed services. We also provide ontology merging process in this section.

### A. Composite Web Services Framework

Figure 2 shows important modules of composite web services framework. Following is the brief description of each module:

*1) Users Subscription:* To provide customer specific composite web services facility, each new user has to register with the system. User Registration module maintains accounts of all users. Each user is given unique identification after registration. Once registered, user can access any of the services listed under CompoServRegistry.

*2) CompoServRegistry:* CompoServRegistry keeps records of all the composite web services which are made available by the framework for registered users. Composite services are listed in the order of frequency of composite services use. The registry is updated each time a new composition is carried out or old service is deleted by the system.

*3) MergOntoBase:* As we have assumed that all services have their OWL-S descriptions, system takes advantage of this fact. For each new composite service, a new OWL-S document is created by merging ontologies of web services participating in a particular service composition. This merged ontology is added to MergOntoBase. MergOntoBase is database containing all the merged ontologies of composite services. Descriptive list of composite services provided by the CompoServRegistry is prepared using these ontologies. If any changes are done in MergOntoBase, same changes get reflected in CompoServRegistry automatically.

*4) Web Service Composer:* The core part of our system is Web Service Composer. It carries out compositions of web services in predefined domains. Composition process is ontology based. It is described in section 3 in detail. After each successful composition, copies of corresponding ontologies are used to generate single merged ontology for all participating web services. Newly generated merged ontology is then added to MergOntoBase.

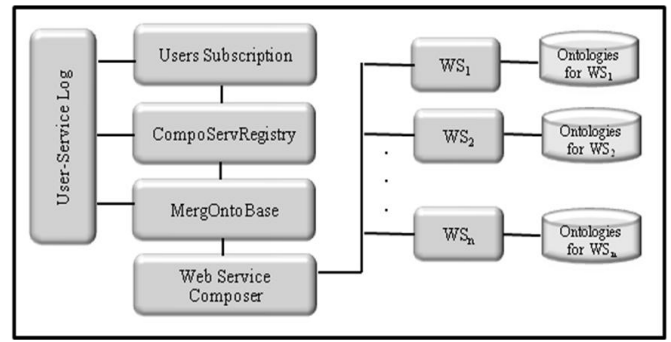

Fig. 2. Composite Web Services Framework

*5)* Framework keeps track of all service composition requests by each user. It stores log of all unique composite service requests for specific user under User-Service Log.

Services listed under CompoServRegistry are sorted using User-Service Log. Frequently used composite services are shown at the top of the list which makes the interface more users friendly.

### B. Maintaining merged ontology repository

Figure three shows how different modules of composite web services interact to maintain MergOntoBase and CompoServRegistry. When a registered user logs into the system, user is provided with list of Composite Services provided by the system i.e. CompoServRegistry. If user selects one of the services from CompoServRegistry, he/she is immediately asked for further inputs. Corresponding composition is carried out and response is sent to user immediately. CompoServRegistry is also updated and one more entry is added in CompoServRegistry.

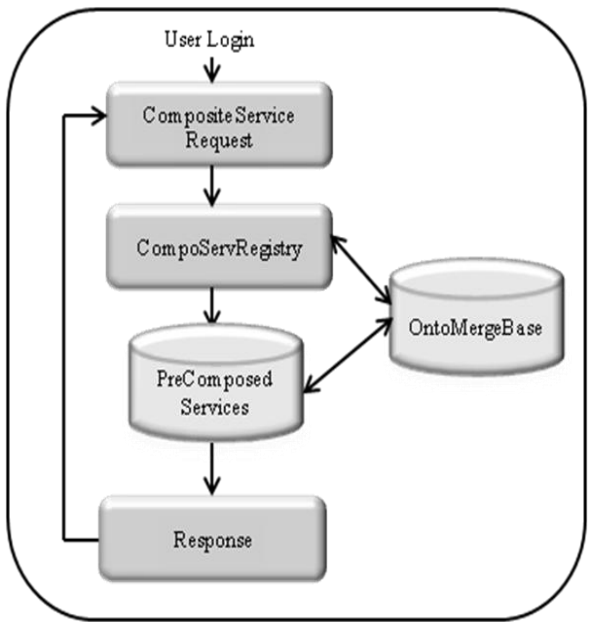

Fig. 3. Updtaing CompServRegistry and OntoMergBase

## V. IMPLEMENTATION OF OUR FRAMEWORK

In this section, we provide description of the system implemented using framework proposed in this paper. We have chosen E-learning domain for implementation. The system provides five composite services under e-learning domain namely

- Learning Resource Library
- Synchronous Learning
- Training Scheduler
- Learning Content Management
- Employee Management

Each of these services is implemented by composing more than three web services. Learning Resource Library provides access to learning resources like slide shows, eBooks and videos. Synchronous Learning composite service provides functions to implement audio and video conferencing, shared white board and chatting. Training scheduler composite service can be used to generate time tables for different subjects, batches and students. Various subject related assignments and completion reports can be maintained automatically by calling services from Learning Content Management composite service. Employee Management composite service provides functions to manage employees in eLearning organization. For example, maintaining personal information, attendance records, maintaining salary records.

### A. Development Environment

As our approach in specifically for web services under semantic web, we have considered all web services having their description in ontology documents. These ontologies are prepared using OWL-S which enables automated composition of web services. It organizes a web service description into four conceptual areas: the process model, the profile, the grounding, and the service. The system is built in Java using Eclipse editor and run on GlassFish Server 3.1. The OWL-S descriptions of web services are built using Protégé Ontology Editor. For completing compositions we use service ontologies using APIs provided by MINDSWAP (Maryland Information and Network Dynamics Lab Semantic Web Agents Project) [15]. For Merging Ontologies we use PROMPT Plug-In of Protégé [14].

### B. Scenario of Learning Resources Library

Out of five composed services provided by the system, we select Learning Resource Library to explain working of the system. Resources like e-books and slide shows are required by any organization running e-learning courses. Under these courses, candidates are offered different subjects. There are number of web sites which allow downloading the resources for these subjects. We consider Learning Resource Library provides functions to access following E-learning resources to registered users:

- EBooks
- Slide shows
- Videos
- Simulations Tools
- Development Tools

Learning Resource Library is prepared by composing five web services each of which provides functions to access the resources under each resource category. We have created ServicePorfile, ServiceModel and ServiceGrounding for each web service used under composition. For completing composition of web services, matching of input, output parameters of functions and their data types in web service is required. For completing web service composition, Service Profiles are used as these contain details of various functions and their input, output parameters for web services. Learning Resource Library aims to provide resources which are listed according to subject names and not according to resource categories. To achieve this, we extract ontologies for all five web services and get subject names for the resources. We then prepare a list of resources which is divided according to subject names. Under each subject name resources are displayed which may be from any of the five resource categories.

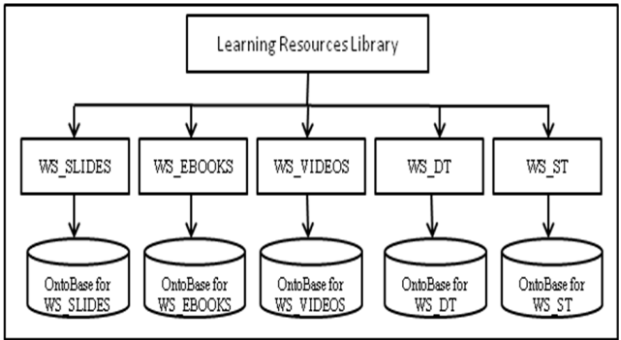

Fig. 4. Service Composition for Learning Resource Library

### C. Ontology Merging Process

To provide description of each composite web service under CompoServRegistry and to reply faster, ontologies of participating web services are merged. There are different types of ontology merging like manual merging, automated merging and semi-automated merging. We adapt semi automated approach. We use Protégé Ontology Editor [12] for unifying source ontologies. PROMPT Plug-In of protégé provides merge option where we select two source ontologies for merging. While merging classes, we consider suggestions provided by PROMPT Plug-In and also provide our suggestions to resolve conflicts. We consider that two classes from different ontologies having similar name but different semantics are not merged. If one class can be mapped to another class, we merge those two classes; otherwise we create a new class. We merge relevant attributes considering differences between data types.

```
<owl:Thing rdf:about="#bk101">
    <rdf:type rdf:resource="#EBOOKS"/>
    <hasAuthor>Gambardella, Matthew</hasAuthor>
    <hasTitle>XML Developer's Guide</hasTitle>
    <hasSubject>Computer</hasSubject>
    <hasPrice>44.95</hasPrice>
    <hasPublish_date>2000-10-01</hasPublish_date>
    <hasDescription>An in-depth look at creating
    applications with XML.</hasDescription>
</owl:Thing>
<owl:Thing rdf:about="#bk102">
    <rdf:type rdf:resource="#EBOOKS"/>
    <hasAuthor>O'Brien, Tim</hasAuthor>
    <hasTitle>Microsoft .NET: The Programming Bible</hasTitle>
    <hasSubject>Computer</hasSubject>
    <hasPrice>36.95</hasPrice>
    <hasPublish_date>2000-12-09</hasPublish_date>
    <hasDescription>Microsoft's .NET initiative is explored in
    detail in this deep programmer's reference.</hasDescription>
</owl:Thing>
<owl:Thing rdf:about="#bk103">
    <rdf:type rdf:resource="#EBOOKS"/>
    <hasAuthor>Jawaharlal Nehru</hasAuthor>
    <hasTitle>The Discovery Of India</hasTitle>
    <hasSubject>History</hasSubject>
    <hasPrice>45.95</hasPrice>
    <hasPublish_date>2000-10-01</hasPublish_date>
    <hasDescription>Gives an understanding of the glorious
    intellectual and spiritual tradition of great country.
    </hasDescription>
</owl:Thing>
```

```
<owl:Thing rdf:about="#slide201">
    <rdf:type rdf:resource="#SLIDES"/>
    <hasAuthor>Doug Tidwell</hasAuthor>
    <hasTitle>Introduction to XML </hasTitle>
    <hasSubject>Computer</hasSubject>
    <hasDescription> History, rules and xml standards are
    explainded</hasDescription>
</owl:Thing>
<owl:Thing rdf:about="#slide202">
    <rdf:type rdf:resource="#SLIDES"/>
    <hasAuthor>Booch, Rumbaugh</hasAuthor>
    <hasTitle>Introduction to UML </hasTitle>
    <hasSubject>Computer</hasSubject>
    <hasDescription> Describes UML notations and UML
    diagrams</hasDescription>
</owl:Thing>
<owl:Thing rdf:about="#slide203">
    <rdf:type rdf:resource="#SLIDES"/>
    <hasAuthor></hasAuthor>
    <hasTitle>History of India</hasTitle>
    <hasSubject>History</hasSubject>
    <hasDescription>Introduces history indian freedom
    strugle</hasDescription>
</owl:Thing>
```

Fig. 6. Part of ontology of WS_SLIDES web service

```
<owl:Thing rdf:about="#s301">
    <rdf:type rdf:resource="#COMPUTER"/>
    <hasEbook>bk101</hasEbook>
    <hasEbook>bk102</hasEbook>
    <hasSlides>slide201</hasSlides>
    <hasSlides>slide202</hasSlides>
</owl:Thing>
<owl:Thing rdf:about="#s302">
    <rdf:type rdf:resource="#HISTORY"/>
    <hasEbook>bk103</hasEbook>
    <hasSlides>slide203</hasSlides>
</owl:Thing>
```

Fig. 7. Part of merged ontology of Learning Resources Library composite service

Figure five, six and seven demonstrate how ontology merging is carried out for Learning Resources Library composite service. Figure five is part of ontology describing ebooks provided by WS_EBOOKS web service for subjects: computer and history. Figure six is part of ontology describing slides available from WS_SLIDES web service for subjects: computer and history. Figure seven is a part of merged ontology used by Learning Resources Library composite service which shows resources categorized according to subject types and not according to resource types. This merged ontology is used by CompoServRegistry to display description of Learning Resources Library

## VI. IMPACT ANALYSIS

To find out effectiveness of our framework we deployed our implemented system. 2500 customers registered to access composite services from the system. We also provided one more portal to provide access to all individual web services. These are the services which are used by the system for generating compositions. We measuring effectiveness, we considered Learning Resources Library composite service. We counted number of resources downloaded by registered users in each month. At the same time, we also maintained count of learning resources downloaded using individual web services. We maintained the count for one year. We observed that there is considerable increase in number of learning resource downloads provided from Learning Resource Library composite service. Figure eight shows statistics of increase in learning downloads for one year.

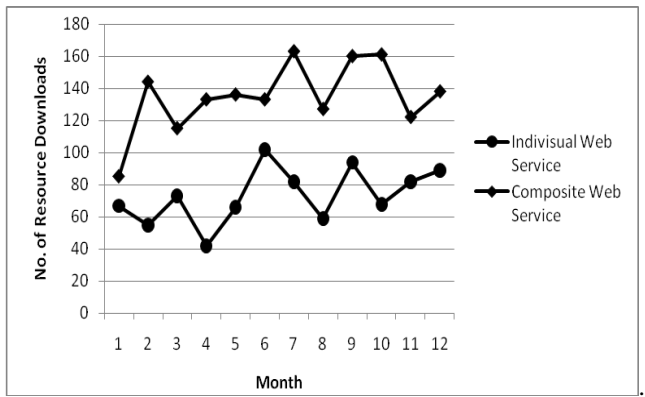

Fig. 8. Effect of precomposed services on resource downloads

We also measured framework effectiveness by counting number of function calls. If any of the precomposed service from implemented system if is called by registered user, all functions under web services used for that composition are called. We count all these function calls. We also count number of calls done for same functions using portal providing individual web services. We maintained this count for three months.

Figure nine show that web service functions are called more number of times using precomposed services than using individual services. These graphs demonstrate how web service providers can grow their business by implementing our proposed framework.

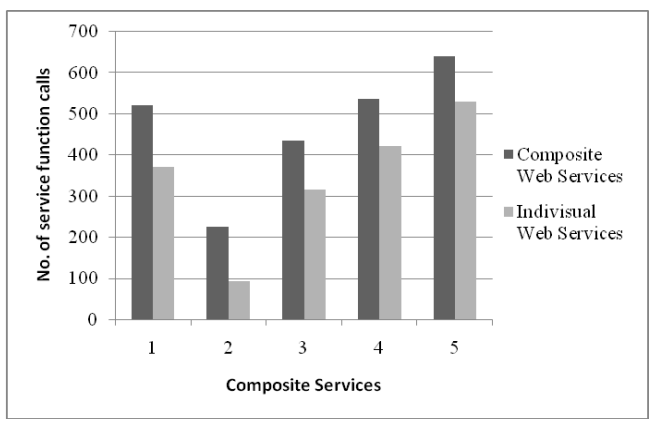

Fig. 9. Effect of precomposed services on service function calls

## VII. CONCLUSION

We have provided a precomposed web services framework which is a semi-automated. The framework is more user friendly than traditional approaches of providing web service compositions. Every time user requests for composite service, user need not to select the required web services explicitly. Instead services will be called by the system implicitly. We evaluated how ontology merging can be effectively used for automated web service composition.

Customer centric composite service system can be implemented using our framework. We further look to design the framework to provide composite web services which will be completely automated.

## REFERENCES

[1] Gexin Li, Shuiguang Deng, Haijiang Xia, Chuan Lin, "Automatic Service Composition Based on Process Ontology", IEEEXplore, 2007

[2] YU Qing-mei, XIAO Peng-yan, JIN Ting, HUANG Dong-mei, "Semantic and Heuristic Approach to the Composition of Web Services," International Conference on Web Information Systems and Mining,, 2010

[3] 3. Jiangang Ma, Yanchun Zhang and Minglu Li., "OMWSC – An Ontology-Based Model for Web Service Composition," Proceedings of Fifth International Conference on Quality Software, IEEEXplore , 2005

[4] 4Yajuan Song, Lei Liu, Ping Ren, "Web Service Composition Based on the Annotated Ontology," Fifth International workshop on education technology and computer science, IEEEXplore, 2009

[5] Liquan Han, Shufen Liu, Lu Han, Zhilin Yao, "Service Composition Engine based on Service-ontology," IET, 2008

[6] Budak Arpinar, Boanerges Aleman-Meza, Ruoyan Zhang and Angela Maduko, "Ontology-Driven Web Service Composition Platform," IEEE International conference on E-Commerce Technology , 2008

[7] Naiwen Lin, Ugur Kuter and James Hendler, "Web Service Composition via Problem Decomposition across Multiple Ontologies," IEEE Congress on Services, 2007

[8] Chao ma, Yanxiang He, Naixue Xiong, Laurence T. Yang, "VFT: An Ontology-based Tool for Visualization and Formalization of Web Service Composition," International Conference on Computational Science and Engineering, IEEEXplore , 2009

[9] Ourania Hatzi, Georgios Meditskos, Dimitris Vrakas, Nick Bassiliades, "Semantic Web Service Composition using Planning and Ontology Concept Relevance," IEEE/WIC/ACM International Joint Conferences on Web Intelligence and Intelligent Agent Technologies, 2009

[10] Fei Tou, Qingxi Hu, Yuan Yao, Gaochun Xu, Minglun Fang, "Study on Web Service Matching and Composition Based on Ontology," WRI World Congress on Computer Science and Information Engineering, 2009

[11] OWL-S: Semantic Markup for Web Services, http://www.w3.org/Submission/OWL-S/#2

[12] Sanjay Kumar Malik, Nupur Prakash, S.A.M. Rizvi, "Ontology Merging using Prompt plug-in of Protege in Semantic Web," International Conference on Computational Intelligence and Communication Networks, IEEEXplore, 2010

[13] Natalya F. Noy and Deborah L. McGuinness, "Ontology Development 101: A Guide to Creating Your First Ontology", Stanford Knowledge Systems Laboratory Technical Report KSL-01-05 and Stanford Medical Informatics Technical Report SMI-2001-0880 , 2001

[14] Protege wiki, http://protegewiki.stanford.edu/wiki/Main_Page

[15] The MINDSWAP group, http://www.mindswap.org/